\pdfoutput=1

\documentclass[11pt]{article}

\usepackage{titlesec}

\titleformat{\paragraph}[runin]
  {\normalfont\normalsize\bfseries}{\theparagraph}{1em}{}[:]

\titlespacing{\paragraph}{0pt}{\parskip}{0.5em}

\usepackage{acl}
\definecolor{mygreen}{RGB}{0,150,0}
\definecolor{myred}{RGB}{255,0,0}
\usepackage{enumitem}
\usepackage{times}
\usepackage[font=small,labelfont=bf]{caption}
\usepackage{array,ragged2e}
\newcolumntype{P}[1]{>{\RaggedRight\arraybackslash}p{#1}}

\definecolor{paired-light-blue}{RGB}{198, 219, 239}
\definecolor{paired-dark-blue}{RGB}{49, 130, 188}
\definecolor{paired-light-orange}{RGB}{251, 208, 162}
\definecolor{paired-dark-orange}{RGB}{230, 85, 12}
\definecolor{paired-light-green}{RGB}{199, 233, 193}
\definecolor{paired-dark-green}{RGB}{49, 163, 83}
\definecolor{paired-light-purple}{RGB}{218, 218, 235}
\definecolor{paired-dark-purple}{RGB}{117, 107, 176}
\definecolor{paired-light-gray}{RGB}{217, 217, 217}
\definecolor{paired-dark-gray}{RGB}{99, 99, 99}
\definecolor{paired-light-pink}{RGB}{222, 158, 214}
\definecolor{paired-dark-pink}{RGB}{123, 65, 115}
\definecolor{paired-light-red}{RGB}{231, 150, 156}
\definecolor{paired-dark-red}{RGB}{131, 60, 56}
\definecolor{paired-light-yellow}{RGB}{231, 204, 149}
\definecolor{paired-dark-yellow}{RGB}{141, 109, 49}

\definecolor{bg1}{HTML}{FF9966}
\definecolor{bg2}{HTML}{CCE5FF}
\definecolor{bg3}{HTML}{FFCC99}
\definecolor{bg4}{HTML}{FFC107}
\definecolor{bg5}{HTML}{FFCCCC}
\definecolor{bg6}{HTML}{D5E8D4}
\definecolor{bg7}{HTML}{eeeeee}
\definecolor{bg8}{HTML}{cdeb8b}
\definecolor{bg9}{HTML}{dae8fc}
\definecolor{bg10}{HTML}{a2e6eb}

\definecolor{bg31}{HTML}{FFCDD2} 

\definecolor{bg32}{HTML}{F8BBD0}

\definecolor{bg33}{HTML}{E1BEE7} 

\definecolor{bg34}{HTML}{D7CCC8} 

\definecolor{bg35}{HTML}{B2DFDB} 

\definecolor{bg36}{HTML}{A5D6A7} 

\definecolor{bg37}{HTML}{FFF9C4} 

\definecolor{bg38}{HTML}{FFECB3} 

\definecolor{bg111}{HTML}{CB6843}

\definecolor{bg112}{HTML}{D77C5C}

\definecolor{bg113}{HTML}{E28E6E}
\definecolor{bg114}{HTML}{E89F7D}
\definecolor{bg115}{HTML}{EDAE8A}
\definecolor{bg116}{HTML}{F0BA95}
\definecolor{bg117}{HTML}{F3C29F}
\definecolor{bg118}{HTML}{F6CCAA}
\definecolor{bg119}{HTML}{F8D5B3}
\definecolor{bg120}{HTML}{FADCBD}
\definecolor{bg121}{HTML}{FCE6C7}

\definecolor{bg39}{HTML}{FFE0B2} 

\definecolor{bg40}{HTML}{3CB371} 

\definecolor{bg43}{HTML}{ffe5d9}

\definecolor{bg15}{HTML}{7FFFD4}

\definecolor{bg17}{HTML}{F0FFFF}

\definecolor{bg18}{HTML}{F5FFFA}

\definecolor{bg19}{HTML}{F8F8FF}

\definecolor{bg20}{HTML}{FFFFFF}

\definecolor{bg21}{HTML}{E1F5FE}

\definecolor{bg22}{HTML}{B3E5FC}

\definecolor{bg23}{HTML}{81D4FA}

\definecolor{bg24}{HTML}{4FC3F7}

\definecolor{bg25}{HTML}{29B6F6}

\definecolor{bg26}{HTML}{03A9F4}

\definecolor{bg27}{HTML}{039BE5}

\definecolor{bg28}{HTML}{0288D1}

\definecolor{bg29}{HTML}{0277BD}

\definecolor{bg30}{HTML}{01579B}

\definecolor{bg16}{HTML}{FFCC99}

\definecolor{pg51}{HTML}{E8F5E9} 
\definecolor{pg52}{HTML}{C8E6C9} 
\definecolor{pg53}{HTML}{B9F6CA} 
\definecolor{pg54}{HTML}{A9DFBF} 
\definecolor{pg55}{HTML}{BCF5A6} 

\definecolor{pg56}{HTML}{BEF1CE} 
\definecolor{pg57}{HTML}{CEF6EC} 
\definecolor{pg58}{HTML}{B7F0B1} 
\definecolor{pg59}{HTML}{B1F2B5} 
\definecolor{pg60}{HTML}{9DF3C4} 

\definecolor{pg61}{HTML}{DEF7E0} 
\definecolor{pg62}{HTML}{E8F8DC} 

\definecolor{pg63}{HTML}{EBF7E7} 
\definecolor{pg64}{HTML}{F0FDF4} 

\definecolor{pg65}{HTML}{F1FEE7} 
\definecolor{pg66}{HTML}{F7FFF6} 
\definecolor{pg67}{HTML}{FCFFE7} 
\definecolor{pg68}{HTML}{F4FFD2} 
\definecolor{pg69}{HTML}{EEFFE2} 
\definecolor{pg70}{HTML}{E3FDF5} 

\definecolor{connect-color}{RGB}{0,0,0}
\definecolor{middle-color}{RGB}{255,255,255}
\definecolor{leaf-color}{RGB}{173,216,230}
\definecolor{line-color}{RGB}{25,25,112}

\usepackage{url}            
\usepackage{booktabs}       
\usepackage{amsfonts}       
\usepackage{nicefrac}       
\usepackage{microtype}      
\usepackage{xcolor}         
\usepackage{amsmath}
\usepackage{graphicx}
\usepackage{enumitem}
\usepackage{multirow}
\usepackage{subcaption}
\usepackage{listings}
\usepackage{tcolorbox}
\usepackage{xspace}
\usepackage{makecell}
\usepackage{soul}               
\setstcolor{red}            
\usepackage{tikz}

\usepackage{longtable}
\usepackage{tablefootnote}
\usepackage[edges]{forest}
\definecolor{hidden-draw}{RGB}{20,68,106}
\definecolor{hidden-pink}{RGB}{255,245,247}
\definecolor{red}{RGB}{255,0,0}

\definecolor{hidden-draw}{RGB}{0,0,0}
\definecolor{hidden-pink}{RGB}{255,182,193}

\usepackage[T1]{fontenc}

\usepackage[utf8]{inputenc}

\usepackage{microtype}

\usepackage{inconsolata}

\usepackage[draft,textsize=footnotesize,textwidth=15mm]{todonotes}

\title{A Comprehensive Survey of Hallucination Mitigation Techniques in Large Language Models}

\newcommand*{\affaddr}[1]{#1}
\newcommand*{\affmark}[1][*]{\textsuperscript{#1}}
\newcommand*{\email}[1]{\texttt{#1}}
\author{
S.M Towhidul Islam Tonmoy\affmark[1], S M Mehedi Zaman\affmark[1], \bf{Vinija Jain\affmark[3,4]}\footnotemark[1], Anku Rani\affmark[2], Vipula Rawte\affmark[2], \\
 \bf{Aman Chadha\affmark[3,4]}\thanks{\,\,\,Work does not relate to position at Amazon.}\,\,, \bf{Amitava Das\affmark[2]}  \\
\affaddr{\affmark[1]Islamic University of Technology, Bangladesh \\
\affmark[2]AI Institute, University of South Carolina, USA \\
\affmark[3]Stanford University, USA, 
\affmark[4]Amazon AI, USA}\\
\email{towhidulislam@iut-dhaka.edu}
}

\begin{document}
\maketitle
\begin{abstract}
As Large Language Models (LLMs) continue to advance in their ability to write human-like text, a key challenge remains around their tendency to ``hallucinate'' -- generating content that appears factual but is ungrounded. This issue of hallucination is arguably the biggest hindrance to safely deploying these powerful LLMs into real-world production systems that impact people's lives \cite{Jain2023HallucinationMitigation}. The journey toward widespread adoption of LLMs in practical settings heavily relies on addressing and mitigating hallucinations. Unlike traditional AI systems focused on limited tasks, LLMs have been exposed to vast amounts of online text data during training. While this allows them to display impressive language fluency, it also means they are capable of extrapolating information from the biases in training data, misinterpreting ambiguous prompts, or modifying the information to align superficially with the input. This becomes hugely alarming when we rely on language generation capabilities for sensitive applications, such as summarizing medical records, customer support conversations, financial analysis reports, and providing erroneous legal advice. Small errors could lead to harm, revealing the LLMs' lack of actual comprehension despite advances in self-learning. This paper presents a comprehensive survey of over thirty-two techniques developed to mitigate hallucination in LLMs. Notable among these are Retrieval-Augmented Generation (RAG) \cite{lewis2021RAG}, Knowledge Retrieval \cite{varshney2023stitch}, CoNLI \cite{lei2023chain}, and CoVe \cite{dhuliawala2023chainofverification}. Furthermore, we introduce a detailed taxonomy categorizing these methods based on various parameters, such as dataset utilization, common tasks, feedback mechanisms, and retriever types. This classification helps distinguish the diverse approaches specifically designed to tackle hallucination issues in LLMs. Additionally, we analyze the challenges and limitations inherent in these techniques, providing a solid foundation for future research in addressing hallucinations and related phenomena within the realm of LLMs.
\end{abstract}

\section{Introduction}
Hallucination in Large Language Models (LLMs) entails the creation of factually erroneous information spanning a multitude of subjects. Given the extensive domain coverage of LLMs, their application extends across numerous scholarly and professional areas. These include, but are not limited to, academic research, programming, creative writing, technical advisement, and the facilitation of skill acquisition. Consequently, LLMs have emerged as an indispensable component in our daily lives, playing a crucial role in dispensing accurate and reliable information. Nevertheless, a fundamental issue with LLMs is their propensity to yield erroneous or fabricated details about real-world subjects. This tendency to furnish incorrect data, commonly referred to as hallucination, poses a significant challenge for researchers in the field. It leads to scenarios where advanced models like GPT-4 and others of its ilk may generate references that are inaccurate or completely unfounded \cite{rawte2023troubling}. This issue arises due to the training phase's pattern generation techniques and the absence of real-time internet updates, contributing to discrepancies in the information output \cite{RAY2023121}.

\tikzset{
  my-box/.style={
    rectangle,
    draw=hidden-draw,
    rounded corners,
    text opacity=1,
    minimum height=1.5em,
    minimum width=40em,
    inner sep=2pt,
    align=center,
    line width=0.8pt,
  },
  leaf/.style={
    my-box,
    minimum height=1.5em,
    text=black,
    align=center,
    font=\normalsize,
    inner xsep=2pt,
    inner ysep=4pt,
    line width=0.8pt,
  }
}

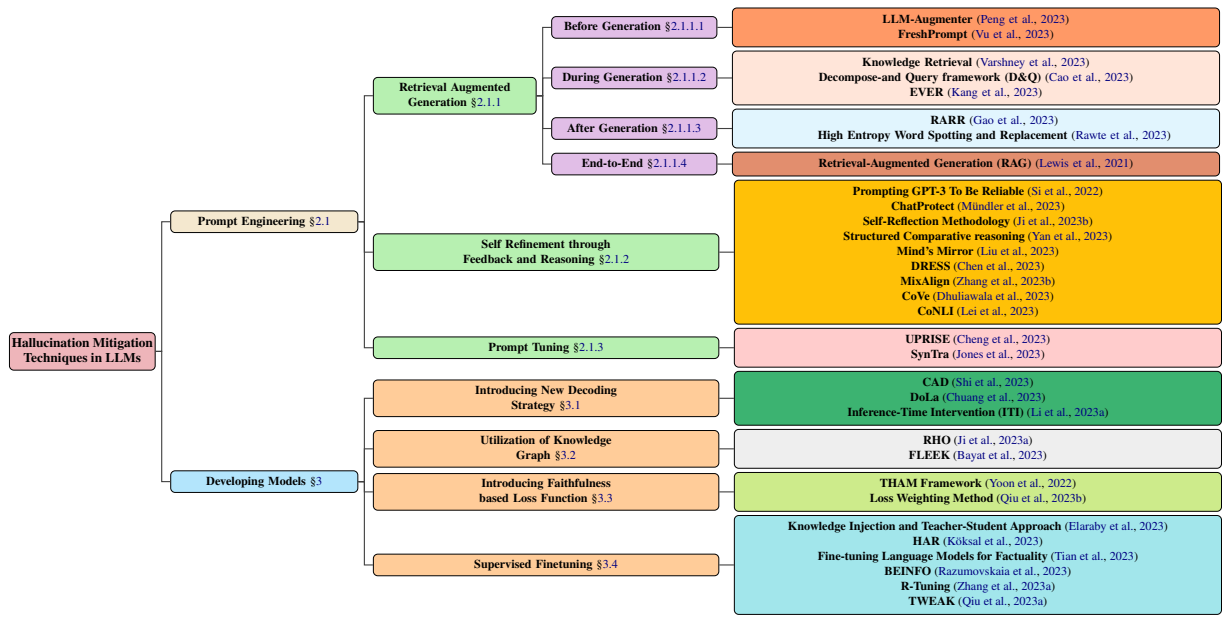
\begin{figure*}[htb!]
  \centering
  \resizebox{\textwidth}{!}{%
    \begin{forest}
      forked edges,
      for tree={
        grow=east,
        reversed=true,
        anchor=base west,
        parent anchor=east,
        child anchor=west,
        base=center,
        font=\large,
        rectangle,
        draw=hidden-draw,
        rounded corners,
        align=center,
        text centered,
        minimum width=5em,
        edge+={darkgray, line width=1pt},
        s sep=3pt,
        inner xsep=2pt,
        inner ysep=3pt,
        line width=0.8pt,
        ver/.style={rotate=90, child anchor=north, parent anchor=south, anchor=center},
      },
      where level=1{text width=15em,font=\normalsize,}{},
      where level=2{text width=14em,font=\normalsize,}{},
      where level=3{minimum width=10em,font=\normalsize,}{},
      where level=4{text width=26em,font=\normalsize,}{},
      where level=5{text width=20em,font=\normalsize,}{},
      [
        \textbf{Hallucination Mitigation}\\ \textbf{Techniques in LLMs}, for tree={fill=paired-light-red!70}
        [
          \textbf{Prompt Engineering} \S\ref{subsec:Prompt Engineering}, for tree={fill=paired-light-yellow!45}
          [
            \textbf{Retrieval Augmented} \\ \textbf{Generation} \S\ref{subsubsec:Retrieval Augmented Generation}, for tree={fill=pg58},text width=13em
            [
            \textbf{Before Generation} \S\ref{subsubsubsec:Before generation}, for tree={fill=bg33},text width=13.2em
            [
            \textbf{LLM-Augmenter} \cite{peng2023check} \\
            \textbf{FreshPrompt} \cite{vu2023freshllms}, for tree={fill=bg1}, leaf
            ]
            ]
            [
            \textbf{During Generation} \S\ref{subsubsubsec:During generation}, for tree={fill=bg33},text width=13.2em
            [
            \textbf{Knowledge Retrieval} \cite{varshney2023stitch} \\
            \textbf{Decompose-and Query framework (D\&Q)} \cite{cao2023step} \\
            \textbf{EVER} \cite{kang2023ever}, for tree={fill=bg43}, leaf
            ]
            ]
            [
            \textbf{After Generation} \S\ref{subsubsubsec:After generation}, for tree={fill=bg33},text width=13.2em
            [
            \textbf{RARR} \cite{gao2023rarr} \\
            \textbf{High Entropy Word Spotting and Replacement} \cite{rawte2023troubling}, for tree={fill=bg21}, leaf
            ]
            ]
            [
            \textbf{End-to-End} \S\ref{subsubsubsec:End-to-End}, for tree={fill=bg33},text width=13.2em
            [
            \textbf{Retrieval-Augmented Generation (RAG)} \cite{lewis2021RAG}, for tree = {fill=bg113}, leaf
            ]
            ]
          ]
          [
            \textbf{Self Refinement through} \\ \textbf{Feedback and Reasoning} \S\ref{subsubsec:Self-refinement through feedback and reasoning}, for tree={fill=pg58},text width=28em
            [
            \textbf{Prompting GPT-3 To Be Reliable} \cite{si2022prompting} \\
            \textbf{ChatProtect} \cite{mündler2023selfcontradictory} \\
            \textbf{Self-Reflection Methodology} \cite{ji2023mitigating} \\
            \textbf{Structured Comparative reasoning} \cite{yan2023basis} \\
            \textbf{Mind’s Mirror} \cite{liu2023minds} \\
            \textbf{DRESS} \cite{chen2023dress} \\
            \textbf{MixAlign} \cite{zhang2023knowledge} \\
            \textbf{CoVe} \cite{dhuliawala2023chainofverification} \\
            \textbf{CoNLI} \cite{lei2023chain}, for tree={fill=bg4}, leaf
            ]
          ]
          [
            \textbf{Prompt Tuning} \S\ref{subsubsec:Prompt Tuning}, for tree={fill=pg58},text width=28em
            [
            \textbf{UPRISE} \cite{cheng2023uprise} \\
            \textbf{SynTra} \cite{jones2023teaching}, for tree={fill=bg5}, leaf
            ]
          ]
        ]
        [
          \textbf{Developing Models} \S\ref{subsec:Developing Models}, for tree={fill=bg22}
          [
            \textbf{Introducing New Decoding} \\ \textbf{Strategy} \S\ref{subsubsec:Introducing new decoding strategy}, for tree={fill=bg16},text width=28em
            [
              \textbf{CAD} \cite{shi2023trusting} \\
              \textbf{DoLa} \cite{chuang2023dola} \\
              \textbf{Inference-Time Intervention (ITI)} \cite{li2023inference}, for tree={fill=bg40}, leaf
            ]
          ]
          [
            \textbf{Utilization of Knowledge} \\ \textbf{Graph} \S\ref{subsubsec:Utilization of Knowledge Graph (KG)},text width=28em, for tree={fill=bg16}
            [
              \textbf{RHO} \cite{ji-etal-2023-rho} \\
              \textbf{FLEEK} \cite{bayat2023fleek}, for tree={fill=bg7}, leaf
            ]
          ]
          [
            \textbf{Introducing Faithfulness} \\ \textbf{based Loss Function} \S\ref{subsubsec:Introducing faithfulness based loss function},text width=28em, for tree={fill=bg16}
            [
              \textbf{THAM Framework} \cite{yoon-etal-2022-information} \\
              \textbf{Loss Weighting Method} \cite{qiu2023detecting}, for tree={fill=bg8}, leaf
            ]
          ]
          [
            \textbf{Supervised Finetuning} \S\ref{subsubsec:Supervised fine-tuning (SFT)},text width=28em, for tree={fill=bg16}
            [
              \textbf{Knowledge Injection and Teacher-Student Approach} \cite{elaraby2023halo} \\
              \textbf{HAR} \cite{köksal2023hallucination} \\
              \textbf{Fine-tuning Language Models for Factuality} \cite{tian2023finetuning} \\
              \textbf{BEINFO} \cite{razumovskaia2023textitdial} \\
              \textbf{R-Tuning} \cite{zhang2023r} \\
              \textbf{TWEAK} \cite{qiu2023think}, for tree={fill=bg10}, leaf
            ]
          ]
        ]
      ]
    \end{forest}
  }
  \caption{Taxonomy of hallucination mitigation techniques in LLMs, focusing on prevalent methods that involve model development and prompting techniques. Model development branches into various approaches, including new decoding strategies, knowledge graph-based optimizations, the addition of novel loss function components, and supervised fine-tuning. Meanwhile, prompt engineering can involve retrieval augmentation-based methods, feedback-based strategies, or prompt tuning.}
  \label{fig:lit_surv}
\end{figure*}

In contemporary computational linguistics, mitigating hallucination is a critical focus. Researchers have proposed various strategies, encompassing feedback mechanisms, external information retrieval, and early refinement in language model generation, to address this challenge. This paper assumes significance by consolidating and organizing these diverse techniques into a comprehensive taxonomy. In essence, the contributions of this paper to the realm of LLM hallucination are threefold:
\vspace{1mm}

\begin{enumerate}
    \item {Introduction of a systematic taxonomy designed to categorize hallucination mitigation techniques for LLMs, encompassing Vision Language Models (VLMs).}
    \item {Synthesis of the essential features characterizing these mitigation techniques, thereby guiding more structured future research endeavors within this domain.}
    \item {Deliberation on the limitations and challenges inherent in these techniques, accompanied by potential solutions and proposed directions for future research.}
\end{enumerate}

\section{Hallucination Mitigation}


The detection of hallucinations has emerged as a significant concern, given the integral role of generative LLMs in critical tasks. \cite{qiu2023detecting} introduced mFACT as a method to identify hallucination in summaries, extending its applicability beyond English to other languages. Additionally, \cite{zhang2023knowledge} proposed a framework for hallucination detection based on contextual information. Another perspective on understanding hallucination causation is presented by \cite{mundler2023self}, who explores self-contradiction as a contributing factor.


\subsection{Prompt Engineering}
\label{subsec:Prompt Engineering}
Prompt engineering is the process of experimenting with various instructions to get the best output possible from an AI text generation model \cite{white2023prompt}. In terms of hallucination mitigation, this process can provide specific context and expected outcomes \cite{feldman2023trapping}. The prompt engineering mitigation techniques can be outlined as follows:
\subsubsection{Retrieval Augmented Generation}
\label{subsubsec:Retrieval Augmented Generation}
Retrieval-Augmented Generation (RAG) enhances the responses of LLMs by tapping into external, authoritative knowledge bases rather than relying on potentially outdated training data or the model's internal knowledge. This approach addresses the key challenges of accuracy and currency in LLM outputs \cite{kang2023ever}. RAG effectively mitigates the issue of hallucination in LLMs by generating responses that are not only pertinent and current but also verifiable, thereby reinforcing user confidence and offering developers an economical way to enhance the fidelity and utility of LLMs across different applications. The mitigation techniques following this system can be further categorized as: 
\paragraph{Before generation} \mbox{} \\
\label{subsubsubsec:Before generation}
For the following techniques, the information retrieval happens before the generation of AI text: \\
\textbf{LLM-Augmenter:}~\cite{peng2023check} proposes a system that augments
a black-box LLM with a set of Plug-And-Play (PnP) \cite{li2023selfchecker} modules. The system makes the LLM generate responses grounded in external knowledge. It
also iteratively revises LLM prompts to improve model responses using feedback generated by utility functions.
In this paper, the authors present LLM-Augmenter to improve LLMs with external knowledge and automated feedback using PnP modules, which do not require any training and can be used instantly. Given a user query, the framework first retrieves evidence from external knowledge and performs reasoning to form evidence chains. Then LLM-Augmenter queries a fixed LLM (GPT-3.5) using a prompt that contains the consolidated evidence for the LLM to generate a candidate response grounded in external knowledge (evidence). LLM-Augmenter then veriﬁes the candidate's response, e.g., by checking whether it
hallucinates evidence. If so, LLM-Augmenter
generates a feedback message. The message is used to revise the prompt to query GPT-3.5 again. The process iterates until a candidate response passes the
veriﬁcation and is sent to the user. \\
\textbf{FreshPrompt:} \cite{vu2023freshllms} address the static nature of most LLMs, highlighting their inability to adapt to the evolving world. The authors introduce FreshQA, a dynamic QA benchmark, evaluating LLMs on questions requiring current world knowledge and those with false premises. Through a two-mode evaluation, correctness and hallucination are measured, revealing limitations and the need for improvement, particularly in fast-changing knowledge scenarios. To address these challenges, the authors present FreshPrompt, a few-shot prompting method that leverages a search engine to incorporate relevant and up-to-date information into prompts. FreshPrompt outperforms competing methods and commercial systems, with further analysis emphasizing the impact of the number and order of retrieved evidence on correctness. The work contributes a detailed evaluation of LLM capabilities in adapting to evolving knowledge, introducing the FreshQA dataset and an effective prompting method, FreshPrompt, to enhance dynamic question answering.
\paragraph{During generation} \mbox{} \\
\label{subsubsubsec:During generation}
The below techniques demonstrate knowledge retrieval at a sentence-by-sentence level, where the model goes through information retrieval while generating each sentence.\\ 
\textbf{Knowledge Retrieval:} \cite{varshney2023stitch} suggest a method that entails actively detecting and reducing hallucinations as they arise. Before moving on to the creation of sentences, the approach first uses the logit output values from the model to identify possible hallucinations, validate that they are accurate, and then mitigate any hallucinations that are found. The most important realization is that handling hallucinations in the generation process is critical because it raises the probability of producing a sentence with hallucinations when the model has previously experienced hallucinations in its output.
This study investigates the use of logit output values -- which are produced by models like the GPT-3 and others -- in the identification of hallucinations. However, it acknowledges that some models available solely through API calls might not give logit output values and emphasizes that this information is a supplementary source rather than a necessary prerequisite for the hallucination detection approach.
The method uses retrieved knowledge as support for the correction phase, instructing the model to repair the phrase by either eliminating or substituting hallucinated information to reduce hallucinations in the created sentence. \\
\textbf{Decompose and Query framework (D\&Q):} In their research, the authors of ~\cite{cao2023step} address challenges faced by LLMs in Question Answering, focusing on hallucinations and difficulties with multi-hop relations. They propose the D\&Q framework to guide models in utilizing external knowledge while constraining reasoning to reliable information, thus mitigating the risk of hallucinations. Experimental results demonstrate D\&Q's effectiveness, showcasing competitive performance against GPT-3.5 on ChitChatQA and achieving a noteworthy 59.6\% F1 score on HotPotQA (question-only). The framework involves a supervised fine-tuning phase without tool invocation, and during the prediction phase, the model uses external tools to query a reliable question-answer base, allowing for backtracking and initiating new searches if needed. The findings underscore D\&Q's potential to enhance the robustness and performance of LLMs in question-answering tasks. \\
\textbf{Real-time Verification and Rectification (EVER):} LLMs often struggle with the challenge of producing inaccurate or hallucinated content, especially in reasoning tasks. In response to this issue prevalent in both non-retrieval-based and retrieval-augmented generation approaches, \cite{kang2023ever} introduces the EVER framework. Unlike existing methods that rectify hallucinations post-hoc, EVER employs a real-time, stepwise strategy during the generation process to detect and rectify hallucinations as they occur. The three-stage process involves generation, validation, and rectification, effectively identifying and correcting intrinsic and extrinsic hallucinations. EVER outperforms both retrieval-based and non-retrieval-based baselines, showcasing significant improvements in generating trustworthy and factually accurate text across diverse tasks such as short-form QA, biography generation, and multi-hop reasoning. The framework's efficacy is empirically validated, demonstrating its ability to mitigate the ``snowballing'' issue of hallucination, making it a valuable contribution to enhancing the accuracy and reliability of LLMs.
\paragraph{After generation} \mbox{} \\
\label{subsubsubsec:After generation}
The following techniques employ the information retrieval system after generating the entirety of its output: \\
\textbf{Retrofit Attribution using Research and Revision (RARR):}~\cite{gao2023rarr} In the realm of LLMs, notable advancements have been achieved across various tasks; however, issues persist, such as generating content without proper support or accuracy. The challenge of determining trustworthiness in LLM outputs, due to a lack of attributability, prompted the introduction of RARR. This model-agnostic system, presented in the introduction, automates the attribution process for any text generation model. Inspired by fact-checking workflows, RARR conducts research and post-editing to align content with retrieved evidence while preserving original qualities, operating seamlessly after LLM generation. Contributions outlined in the introduction encompass formalizing the Editing for Attribution task, introducing new metrics, benchmarking existing revision models, and proposing a research-and-revise model. The conclusion underscores RARR's ability to enhance attribution while preserving essential text properties, providing a practical solution to bolster the reliability of LLM outputs. \\
\textbf{High Entropy Word Spotting and Replacement:} While the technical feasibility of detecting high entropy words may be apparent, a significant challenge arises due to the closed-source nature of many contemporary LLMs, with subscription-based APIs limiting accessibility. The proposed solution by \cite{rawte2023troubling} involves utilizing open-source LLMs to identify high entropy words, followed by their replacement using a lower Hallucination Vulnerability Index-based LLM. The results underscore the exceptional performance of albert-large-v2 \cite{Lan2020ALBERT:} in detecting high entropy words in GPT-3-generated content. Conversely, distilroberta-base \cite{sanh2019distilbert} exhibits superior performance in replacing high entropy words, leading to a reduction in hallucinations. An integral aspect of this approach is the treatment of consecutive high-entropy words as a unified unit, where these words are collectively masked before replacement, proving particularly effective in addressing hallucinations related to Generated Golem or Acronym Ambiguity.
\paragraph{End-to-End RAG} \mbox{} \\
\label{subsubsubsec:End-to-End}
The end-to-end process of RAG proposed in the paper by \cite{lewis2021RAG} involves integrating a pre-trained sequence-to-sequence (seq2seq) transformer with a dense vector index of Wikipedia, accessed through the Dense Passage Retriever (DPR). This innovative combination allows the model to condition its output generation on both the input query and latent documents provided by the DPR.

In this process, the DPR acts as a neural retriever, supplying relevant documents based on the input. These documents are then used by the seq2seq model, specifically BART, to generate the final output. The model employs a top-K approximation to marginalize these latent documents, which can be done on a per-output basis (assuming one document is responsible for all tokens) or a per-token basis (allowing different documents to influence different parts of the output). 

Crucially, both the generator and the retriever in this RAG setup are trained end-to-end, ensuring that they learn jointly and improve each other's performance. This methodology contrasts with previous approaches that required architectures with non-parametric memory to be built from scratch for specific tasks. Instead, RAG uses pre-trained components, pre-loaded with extensive knowledge, allowing the model to access and integrate a vast range of information without the need for additional training. This end-to-end approach results in enhanced performance on various knowledge-intensive tasks, demonstrating the efficacy of combining parametric and non-parametric memory in generation models.

\subsubsection{Self-refinement through feedback and reasoning}
\label{subsubsec:Self-refinement through feedback and reasoning}
After an LLM provides an output for a specific prompt, proper feedback about the output can make the LLM give better and more accurate outputs in its consecutive iterations \cite{madaan2023selfrefine}. Abiding by this method, the following are the specific hallucination mitigation techniques: \\
\textbf{Prompting GPT-3 To Be Reliable:} According to \cite{si2022prompting}'s paper, LLMs, particularly GPT-3, exhibit remarkable few-shot prompting abilities, enhancing their applications in real-world language tasks. Despite this, the issue of improving GPT-3's reliability remains underexplored. This study decomposes reliability into four crucial facets -- generalizability, social biases, calibration, and factuality -- and introduces simple and effective prompts to enhance each aspect. The research surpasses smaller-scale supervised models on all reliability metrics, offering practical strategies for improving GPT-3's performance. The paper outlines previous works on LLM reliability, highlighting the novelty of this study's comprehensive analysis and focus on effective prompting strategies. Drawing inspiration from ML safety surveys, the reliability framework aligns with identified risks in existing conceptual frameworks. Lastly, the systematic exploration of GPT-3's reliability has been summarized, which introduces practical prompting strategies, and emphasizes the study's contribution to insights into LLMs and practical recommendations for GPT-3 users. \\
\textbf{ChatProtect:}~\cite{mündler2023selfcontradictory} focuses on an important type of
hallucination called self-contradiction, which occurs when an LLM generates two logically inconsistent sentences given the same context.
They propose a three-step pipeline for reasoning about self-contradictions. Importantly, the approach is built upon prompting strategies, making it applicable to black-box LLMs without requiring external grounded knowledge. They conducted an extensive evaluation targeting four modern instruction-tuned LMs on the task of open-domain text generation, demonstrating the substantial benefits of the approach: it effectively exposes self-contradictions, accurately detects them, and appropriately mitigates their occurrence. \\
\textbf{Self-Reflection Methodology:} The paper \cite{ji2023mitigating} explores and addresses the phenomenon of hallucination in medical generative QA systems utilizing widely adopted LLMs and datasets. The focus is on identifying and understanding problematic answers, emphasizing hallucination.
To tackle this challenge, the paper introduces an interactive self-reflection methodology that integrates knowledge acquisition and answer generation. Through this iterative feedback process, the approach systematically improves the factuality, consistency, and entailment of generated answers. Leveraging the interactivity and multitasking ability of LLMs, the method produces progressively more precise and accurate answers. Experimental results, both automatic and human evaluations, highlight the effectiveness of this approach in reducing hallucinations compared to baselines.
The investigation into hallucinations in generation tasks, particularly in the medical domain, is crucial for AI's accountability and trustworthiness. The proposed iterative self-reflection method, employing a generate-score-refine strategy on background knowledge and answers, is empirically proven to be effective, generalizable, and scalable in mitigating hallucinations. \\
\textbf{Structured Comparative (SC) reasoning:} In the realm of text preference prediction, where LLMs often grapple with inconsistencies in reasoning, \cite{yan2023basis} introduces the SC reasoning method. SC employs a prompting approach that predicts text preferences by generating structured intermediate comparisons. It starts by proposing aspects of comparison and then generates textual comparisons under each aspect. Utilizing a pairwise consistency comparator, SC ensures that each aspect's comparisons distinctly differentiate between texts, effectively reducing hallucination and enhancing consistency. The methodology is showcased across various NLP tasks, including summarization, retrieval, and automatic rating, demonstrating that SC equips LLMs with state-of-the-art performance in text preference prediction. The structured reasoning approach of SC, along with its consistency enforcement, is validated through comprehensive evaluations and ablation studies, emphasizing its effectiveness in improving accuracy and coherence across diverse tasks. Human evaluations further underscore SC's interpretative capabilities, assisting users in making informed decisions. \\
\textbf{Mind’s Mirror:} While chain-of-thought (CoT) distillation methods show promise for downsizing LLMs to small language models (SLMs), there is a risk of carrying over flawed reasoning and hallucinations. To address this, \cite{liu2023minds} proposed a methodology with two key components: First, a novel approach introduces distilling the self-evaluation capability inherent in LLMs into SLMs, aiming to mitigate adverse effects and reduce hallucinations. Second, a comprehensive distillation process incorporates multiple distinct CoT and self-evaluation paradigms for holistic knowledge transfer into SLMs.

The methodology trains SLMs to possess self-evaluation capabilities, recognizing and correcting hallucinations and unreliable reasoning, enhancing predictive accuracy and reliability on various NLP tasks. Comprehensive experiments demonstrate the superiority of this method across reasoning tasks, offering a promising approach to responsibly downsize LLMs. \\
\textbf{DRESS:}~\cite{chen2023dress} propose using natural language feedback (NLF), specifically critique and refinement NLF, to improve alignment with human preferences and interaction capabilities of large vision language models (LVLMs). They generalize conditional reinforcement learning to effectively incorporate non-differentiable NLF by training the model to generate corresponding responses conditioned on the NLF. Experiments show relative improvements in DRESS over prior state-of-the-art LVLMs in metrics of helpfulness, honesty, and harmlessness alignment. \\
\textbf{MixAlign:} Despite having accurate reference points, LLMs may disregard them and rely on incorrect references or biases instead. This tendency to hallucinate arises when users ask questions that do not directly align with the retrieved references, lacking detailed knowledge of the stored information. \cite{zhang2023knowledge} focus on this knowledge alignment problem and introduce MixAlign, a framework that interacts with both the user and knowledge base to clarify how the user question relates to the stored information. MixAlign uses a language model to achieve automatic knowledge alignment and, if needed, further enhances this alignment through user clarifications. MixAlign focuses on utilizing grounding knowledge for faithful decision-making. In cases of uncertainty or unclear evidence, MixAlign generates a question seeking clarification from the user - a process referred to as human-assisted knowledge alignment. \\
\textbf{Chain-of-Verification (CoVe):}~\cite{dhuliawala2023chainofverification} develop the CoVe method where the model 
\begin{enumerate}
\item Drafts an initial response.
\item Plans verification questions to fact-check its draft.
\item Answers those questions independently so the answers are unbiased.
\item Generates a final verified response.
\end{enumerate}
 
Experiments show CoVe decreases hallucinations across tasks like list-based Wikidata questions and long-form text generation. Given a user query, an LLM generates a baseline response that may contain inaccuracies like factual hallucinations. CoVe first generates verification questions to ask, then answers them to check for agreement.
 \\
\textbf{Chain of Natural Language Inference (CoNLI):} ~\cite{lei2023chain} address the challenge of hallucinations generated by LLMs when provided background context. Despite fluency in natural language generation, LLMs often produce ungrounded hallucinations unsupported by the given sources.

The proposed hierarchical framework focuses on detecting and mitigating such hallucinations without requiring fine-tuning or domain-specific prompts. The framework utilizes Chain of Natural Language Inference (CoNLI) for state-of-the-art hallucination detection by identifying ungrounded content. Post-editing is then used to reduce hallucinations and enhance text quality without model adjustment. Extensive experiments on text-to-text datasets demonstrate effectiveness in both hallucination detection and reduction. By formulating detection as a chain of natural language inference tasks, the framework incorporates sentence and entity-level judgments with interpretability.

The plug-and-play framework allows seamless deployment across contexts with competitive hallucination detection and reduction performance while preserving text quality. 

\subsubsection{Prompt Tuning}
\label{subsubsec:Prompt Tuning}
Prompt tuning is a technique that involves adjusting the instructions provided to a pre-trained LLM during the fine-tuning phase to make the model more effective at specific tasks. The LLM learns from `Soft Prompts', which are not predetermined but are instead learned by the model through backpropagation during the fine-tuning \cite{lester-etal-2021-power}. For hallucination mitigation, the following techniques, which involve prompt tuning, have been proposed as of now: \\  
\textbf{Universal Prompt Retrieval for Improving zero-Shot Evaluation (UPRISE):}~\cite{cheng2023uprise} propose UPRISE, which tunes a lightweight and versatile retriever that automatically retrieves prompts for a given zero-shot task input. Specifically, they demonstrate universality in a cross-task and cross-model scenario: the retriever is tuned on a diverse set of tasks, but tested on unseen type tasks. The retriever is trained to retrieve prompts for multiple tasks, enabling it to generalize to unseen task types during inference. \\
\textbf{SynTra:} Large language models (LLMs) often exhibit hallucination in abstractive summarization tasks, even when the necessary information is present. Addressing this challenge is difficult due to the intricate evaluation of hallucination during optimization. \cite{jones2023teaching} introduce SynTra, a method that uses a synthetic task to efficiently reduce hallucination on downstream summarization tasks. SynTra optimizes the LLM's system message via prefix-tuning on the synthetic task, then transfers this capability to more challenging, realistic summarization tasks. Experiments demonstrate reduced hallucination for two 13B parameter LLMs, highlighting the effectiveness of synthetic data for mitigating undesired behaviors.

\section{Developing Models}
\label{subsec:Developing Models}
Some papers focused on developing novel models to mitigate hallucinations. It is an ongoing and evolving process requiring a combination of algorithmic advancements and data quality improvements. Instead of going for fine-tuning models, the following techniques implemented whole model architecture to tackle hallucinations. These techniques can be categorized as follows:
\subsection{Introducing new decoding strategy}
\label{subsubsec:Introducing new decoding strategy}
Decoding strategy generally involves designing techniques that specifically target the generation phase of a model. In terms of hallucination, the techniques aim to reduce the occurrence of hallucinations in the generated outputs by guiding the generation phase towards authentic or context-specific generation \cite{lango-dusek-2023-critic}. The following techniques make use of the decoding strategy:

\textbf{Context-Aware Decoding (CAD):}~\cite{shi2023trusting} present CAD, which follows a contrastive output distribution that amplifies the difference between the output probabilities when a model is used with and without context.
CAD is particularly effective in overriding a model’s prior
knowledge when it contradicts the provided
context, leading to substantial improvements in
tasks where resolving the knowledge conflict is
essential. CAD can be used with off-the-shelf pre-trained language models without any additional training. More notably, CAD is especially beneficial for knowledge-conflicting tasks, where the context contains information contradictory to the model’s
prior knowledge. The results demonstrate the potential of
CAD in mitigating hallucinations in text generation
and overriding prior knowledge with reliable and
trusted information. \\
\textbf{Decoding by Contrasting Layers (DoLa):} \cite{chuang2023dola} introduce DoLa, a simple decoding strategy designed to mitigate hallucinations in pre-trained LLMs without the need for external knowledge conditioning or additional fine-tuning.
DoLa achieves the next-token distribution by contrasting logit differences between later and earlier layers projected into the vocabulary space. This leverages the observed localization of factual knowledge in specific transformer layers. Consequently, DoLa enhances the identification of factual knowledge and minimizes the generation of incorrect facts. Across various tasks, including multiple-choice and open-ended generation tasks like TruthfulQA, DoLa consistently improves truthfulness, enhancing the performance of LLaMA family models. \\
\textbf{Inference-Time Intervention (ITI):}~\cite{li2023inference} introduce ITI, a technique designed to enhance the “truthfulness” of LLMs. ITI operates by shifting model
activations during inference, following a set of directions across a limited number of attention heads. This intervention significantly improves the performance of LLaMA models on the TruthfulQA benchmark. The technique first identifies a sparse set of attention heads with high linear probing accuracy for truthfulness. Then, during inference, they shift activations along these
truth-correlated directions. It repeats the same intervention autoregressively until the whole answer is generated. ITI results in a significant performance increase on the TruthfulQA benchmark.

\subsection{Utilization of Knowledge Graph (KG)}
\label{subsubsec:Utilization of Knowledge Graph (KG)}
KGs are organized collections of data that include details about entities (i.e., people, places, or objects), their characteristics, and the connections between them \cite{sun2023head}. It arranges data such that machines can comprehend the relationships and semantic meaning of the material. KGs offer a basis for sophisticated reasoning, data analysis, and information retrieval. Thus, several studies have used KGs in the context of hallucination mitigation \cite{bayat2023fleek}. They are:

\textbf{RHO:} To handle the hallucination challenge in dialogue response generation, \cite{ji-etal-2023-rho} proposes a framework called RHO that utilizes the representations of linked entities and relation predicates from a KG to generate more faithful responses. To improve faithfulness, they introduce local and global knowledge-grounding techniques into dialogue generation and further utilize a conversational reasoning model to re-rank the generated responses. These two knowledge groundings help the model effectively encode and inject the knowledge information from context-related subgraphs with proper attention. Their work improves the fusion and interaction between external knowledge and dialogue context via various knowledge groundings and reasoning
techniques, further reducing hallucination.

\noindent
\textbf{FactuaL Error detection and correction with Evidence Retrieved from external Knowledge (FLEEK):} \cite{bayat2023fleek} introduce FLEEK, an intelligent and model-agnostic tool aimed at aiding end users, such as human graders, in fact verification and correction. FLEEK features a user-friendly interface capable of autonomously identifying potentially verifiable facts within the input text. It formulates questions for each fact and queries both curated knowledge graphs and the open web to gather evidence. The tool subsequently verifies the correctness of the facts using the acquired evidence and proposes revisions to the original text. The verification process is inherently interpretable, with extracted facts, generated questions, and retrieved evidence directly reflecting the information units contributing to the verification process. For instance, FLEEK would visually highlight verifiable facts with distinct colors indicating their factuality levels, allowing users to interact with clickable highlights that reveal evidence supporting or refuting each claim. Future work includes comprehensive evaluations of FLEEK, testing its compatibility with various LLMs, and subjecting it to a comprehensive benchmark.

\subsection{Introducing faithfulness based loss function}
\label{subsubsec:Introducing faithfulness based loss function}
Creating a metric to gauge how closely a model's outputs match input data or ground truth is the task of this section. In this sense, faithfulness describes the model's capacity to faithfully and properly reflect data from the input without adding errors, omissions, or distortions \cite{chrysostomou2021enjoy}. The following methods portray the use of technique: 

\textbf{Text Hallucination Mitigating (THAM) Framework:}~\cite{yoon-etal-2022-information} introduce the THAM framework for Video-grounded Dialogue. THAM considers the text hallucination problem, which copies
input texts for answer generation without the understanding of the question. It mitigates feature-level hallucination effects
by introducing information-theoretic regularization. THAM framework incorporates Text Hallucination Regularization (THR) loss derived from the mutual information between the response language model and the proposed hallucination language model. Minimizing THR loss contributes to reducing indiscriminate text copying and boosting dialogue performances.
THAM framework incorporates Text Hallucination Regularization loss derived from the proposed information-theoretic text
hallucination measurement approach. \\
\textbf{Loss Weighting Method:}~\cite{qiu2023detecting} focus on low resource language summarization and develops a novel metric, mFACT to evaluate the faithfulness of non-English summaries, leveraging translation-based transfer from multiple English faithfulness metrics. It is developed from four English faithfulness metrics. They study hallucination in a cross-lingual transfer setting. They apply mFACT to study the faithfulness in summarisation of the recent multilingual LLMs. The proposed metric consists of weighting
training samples’ loss based on their faithfulness
score. The experiments show that while common cross-lingual transfer methods benefit summarisation performance, they amplify hallucinations compared to monolingual counterparts. To reduce these hallucinations, they adapt several monolingual methods to cross-lingual transfer and propose a new method based on
weighting the loss according to the mFACT score of each training example.

\subsection{Supervised fine-tuning (SFT)}
\label{subsubsec:Supervised fine-tuning (SFT)}
SFT  serves as a vital phase in aligning LLMs for downstream tasks using labeled data. It helps the model follow human commands for specific tasks \cite{wang2023selfinstruct,chung2022scaling,iyer2023optiml,sun2023principledriven} and eventually increases the faithfulness of the model's outputs. In the context of SFT, the quality of the data stands as the most pivotal concern, as it directly determines the fine-tuned model's performance\cite{xu2023wizardlm,touvron2023llama}. During supervised fine-tuning, the LLM's weights are adjusted based on the gradients from a task-specific loss function that measures the difference between the LLM's predictions and ground truth labels. This technique has proven particularly effective in enhancing the adaptability of LLMs, enabling them to excel at previously unseen tasks.

\textbf{Knowledge Injection and Teacher-Student Approaches:}~\cite{elaraby2023halo} focus on measuring and reducing hallucinations in weaker open-source large language models (LLMs) like BLOOM 7B \cite{workshop2022bloom}. They introduce HALOCHECK, a lightweight knowledge-free framework to quantify hallucination severity in LLMs. The authors explore techniques like knowledge injection and teacher-student approaches to alleviate hallucinations in low-parameter LLMs. The framework uses sentence-level entailment to quantitatively assess hallucination levels.

The work aims to enhance smaller LLM knowledge through Knowledge Injection (KI) by fine-tuning with domain knowledge, without relying on expensive instructions from stronger models. They investigate leveraging a more powerful LLM like GPT-4 to guide weaker LLMs by generating detailed question answers. By assessing hallucination severity, they optimize teacher LLM engagement to reduce the computational costs of relying extensively on large models. This alleviates the need for frequent queries to the teacher model. \\
\textbf{Hallucination Augmented Recitations (HAR):}~\cite{köksal2023hallucination} introduce the concept of attribution in LLMs to control information sources and enhance factuality. While existing methods rely on open-book question answering to improve attribution, the challenge arises when factual datasets reward models for recalling pretraining data rather than demonstrating true attribution. To address this, the authors propose HAR, a novel approach utilizing LLM hallucination to create counterfactual datasets and enhance attribution. Through a case study on open book QA, specifically CF-TriviaQA, the results demonstrate that models fine-tuned with these counterfactual datasets significantly improve text grounding and outperform those trained on factual datasets, even with smaller training datasets and model sizes. The observed improvements are consistent across various open-book QA tasks, including multi-hop, biomedical, and adversarial questions. \\
\textbf{Fine-tuning Language Models for Factuality:} \cite{tian2023finetuning} address hallucination by leveraging recent NLP innovations, employing automated fact-checking methods and preference-based learning through the Direct Preference Optimization algorithm. The researchers fine-tune the Llama-2 model for factuality without human labeling, achieving notable error reductions, particularly in biographies and medical questions. Their approach involves reference-based and reference-free truthfulness evaluations, demonstrating a cost-effective way to enhance model factuality in long-form text generation. The study proposes new benchmark tasks, discusses future avenues, and highlights the potential scalability of factual reinforcement learning for larger models in safety-critical domains. \\
\textbf{BEINFO:} To mitigate the issue and increase
faithfulness of information-seeking dialogue
systems, \cite{razumovskaia2023textitdial} introduce BEINFO, a simple yet effective method that applies behavioral tuning to aid information-seeking dialogue. In this work, the authors propose BEINFO, a simple yet effective method that applies ‘behavioral finetuning’ to increase the faithfulness of the generated responses for information-seeking dialogue. The model is tuned on a large collection of dialogues with the true knowledge source(s) extended with randomly sampled facts from a large knowledge
base. \\
\textbf{Refusal-Aware Instruction Tuning (R-Tuning):} In their recent work, ~\cite{zhang2023r} present a novel approach called R-Tuning for instilling refusal skills in large language models (LLMs). This approach formalizes the idea of identifying knowledge gaps between an LLM's parametric knowledge and the instructional tuning data used to train it. Based on this knowledge gap, R-Tuning constructs refusal-aware training data to teach the LLM when to refrain from responding, specifically when a question falls outside its competence.
The R-Tuning methodology involves two key steps:
\begin{enumerate}
    \item Measuring the knowledge gap between the LLM's parametric knowledge and the instructional tuning questions, to identify uncertain questions. By inferring on the training data once and comparing predictions to labels, the tuning data is separated into uncertain questions and certain questions. 
    \item Constructing refusal-aware training data by appending refusal expressions to uncertain training examples, before fine-tuning the LLM on this data.
\end{enumerate}
\textbf{Think While Effectively Articulating Knowledge (TWEAK):} To reduce hallucinations, \cite{qiu2023think} propose a new decoding method called TWEAK. The method treats the generated sequences at each step and their future sequences as hypotheses. It ranks each generation candidate based on how well their corresponding hypotheses support the input facts, using a Hypothesis Verification Model (HVM). 

The authors tweak only the decoding process without retraining the generative models. This makes their approach easily integrated with any knowledge-to-text generator. Existing decoding methods like beam search sample candidates only based on predicted likelihood, without considering faithfulness. The authors propose a new dataset called FATE, which aligns input facts with original and counterfactual descriptions at the word level.


\section{Conclusion}
This survey paper delves into the critical issue of hallucination in LLMs, emphasizing the widespread impact of LLMs across various domains in our lives. The paper highlights the challenge posed by LLMs generating incorrect information and identifies it as a significant concern for researchers working on prominent LLMs like GPT-4.
The paper explores recent advancements in the detection of hallucinations, with methods such as mFACT, contextual information-based frameworks, and the investigation of self-contradiction as a contributing factor. It underscores the importance of addressing hallucination in LLMs due to their integral role in critical tasks.
The central contribution of the paper lies in presenting a systematic taxonomy for categorizing hallucination mitigation techniques in LLMs, extending its coverage to VLMs. By synthesizing essential features characterizing these techniques, the paper provides a foundation for more structured future research within the domain of hallucination mitigation. Additionally, the paper deliberates on the inherent limitations and challenges associated with these techniques, proposing directions for future research in this area.

In essence, this survey paper not only sheds light on the gravity of hallucination in LLMs but also consolidates and organizes diverse mitigation techniques, contributing to the advancement of knowledge in the field of computational linguistics. It serves as a valuable resource for researchers and practitioners seeking a comprehensive understanding of the current landscape of hallucination in LLMs and the strategies employed to address this pressing issue.

\section{Discussion and Limitations}
Hallucination mitigation in LLMs represents a multifaceted challenge addressed through a spectrum of innovative techniques. The methodologies discussed, ranging from post-generation refinement to supervised fine-tuning, underscore the gravity of the hallucination issue and the pressing need for comprehensive solutions.

In the realm of post-generation refinement, RARR stands out, automating the attribution process and aligning content with retrieved evidence. High Entropy Word Spotting and Replacement tackles hallucinations induced by high-entropy words in LLM-generated content, showcasing the significance of context-aware replacements.

Self-refinement through feedback and reasoning brings forth impactful strategies like ChatProtect, focusing on self-contradiction detection, and Self-Reflection Methodology, employing an iterative feedback process for hallucination reduction in medical generative QA systems. Structured Comparative reasoning introduces a structured approach to text preference prediction, enhancing coherence and reducing hallucination.

Prompt tuning emerges as a powerful technique, with innovations like UPRISE demonstrating the versatility of prompt-based adjustments. SynTra introduces synthetic tasks for mitigating hallucinations in abstractive summarization, offering scalability but raising questions about effectiveness compared to human feedback.

The development of novel models emphasizes decoding strategies such as CAD and DoLa, both instrumental in reducing hallucinations by guiding the generation phase. KG utilization and faithfulness-based loss functions also play crucial roles, as seen in methods like RHO and THAM Framework.

Supervised fine-tuning, a pivotal phase, is explored through various lenses, such as Knowledge Injection and Teacher-Student Approaches, where domain-specific knowledge is injected into weaker LLMs and approaches like HAR employ counterfactual datasets for improved factuality.

Future developments and improvements in a variety of areas are anticipated for language models' approach to hallucination mitigation. The creation of hybrid models, which offer a thorough defense against hallucinations by seamlessly integrating numerous mitigation approaches, is one important direction. By reducing reliance on labeled data, investigating the possibilities of unsupervised or weakly supervised learning techniques might improve scalability and flexibility. In addition, it will be essential to look into the moral ramifications and societal effects of hallucination mitigation strategies to guarantee responsible implementation and promote user confidence. Research on designs specifically intended to reduce hallucinations is further encouraged by the changing field of LLMs, which could lead to the development of new models with built-in safety features. It will be crucial for researchers, business professionals, and ethicists to work together continuously to improve methods, benchmark models, and set standards that put user comprehension and authenticity first. The building of language models that produce coherent and contextually relevant information while simultaneously demonstrating heightened awareness and mitigation of hallucinatory outputs is the field's collective goal as it navigates these future possibilities.

The collected works on hallucination mitigation reveal a diverse array of strategies, each contributing uniquely to address the nuances of hallucination in LLMs. As the field evolves, the synthesis of these approaches could pave the way for more robust and universally applicable solutions, fostering trust and reliability in language generation systems.

Finally, the division of the mitigation techniques surveyed can be easily comprehensible through table \ref{tab:table_1}. 


\bibliography{anthology, custom}

\appendix

\captionsetup{width=14.5cm}
\tiny
\clearpage
\onecolumn
\renewcommand{\arraystretch}{1.8}
\begin{longtable}
{|P{0.35in}|P{0.5in}|P{0.7in}|P{0.3in}|P{0.5in}|P{0.6in}|P{0.5in}|P{0.5in}|P{0.55in}|}
    \caption{Summary of all the works related to hallucination mitigation in two categories. Here, we have divided each \newline work by the following factors:1. Mitigation Technique, 2. Detection, 3. Task(s), 4. Metrics, and 5. Evaluated LLM(s), \newline 6. Dataset(s). {\color{mygreen}\checkmark} indicates that it is present in the paper whereas {\color{myred}$\times$} indicates it is not present.}\label{tab:summary} \\ 
    \hline 
    \raggedbottom \textbf{Category}&
    \textbf{Mitigation Technique(s)}&
    \textbf{Title}&
    \textbf{Detection}&
    \textbf{Task(s)}&
    \textbf{Metric(s)}&
    \textbf{Evaluated LLM(s)}&
    \textbf{Dataset(s)}&
    \textbf{Limitation(s)}\\
    \hline
    \endfirsthead
    \multicolumn{9}{c}%
    {\tablename\ \thetable\ -- \textit{Continued from the previous page}} \\
    \hline
    \textbf{Category} &  \textbf{Mitigation Technique(s)}    &   \textbf{Title} &   \textbf{Detection} & \textbf{Task(s)} & \textbf{Metric(s)} & \textbf{Evaluated LLM(s)}& \textbf{Dataset(s)} & \textbf{Limitation(s)}\\
    \hline
    \endhead
    \hline \multicolumn{5}{r}{\textit{Continued on the next page}} \\
    \endfoot 
    \hline
    \endlastfoot

\multirow{14}{*}[-30ex]\textbf{Prompt \newline Engineering} &
\centering Retrieval Augmented Generation (Before Generation) &
\raggedright Check Your Facts and Try Again: Improving Large Language Models with External Knowledge and Automated Feedback \cite{peng2023check}&
{\color{myred}$\times$}&
\raggedright Information seeking dialog and open-domain Wiki question Answering&
\raggedright KF1, BLEU-4, ROUGE-1, \newline METEOR,
BLEURT, \newline BERTScore, \newline chrF, \newline BARTScore &GPT-3.5&Manual,\newline OTT-QA 
& $\bullet$ Interactive feedback with ChatGPT slows down user experience as it requires multiple queries per response.
$\bullet$ No human evaluation of responses has been conducted yet.
\\
\cline{3-9}
& & \raggedright FRESHLLMS: Refreshing Large Language Models with Search Engine Augmentation \cite{vu2023freshllms} & {\color{mygreen}\checkmark}&QA & Accuracy & T5, Palm, Palmchilla, Flan-T5, Flan-Palm, GPT-3.5, Codex, GPT-4 & FreshQA (Own dataset)
& $\bullet$ Answers can become stale between manual updates by maintainers. 

$\bullet$ Method relies on Google search API, simple English questions, and in-context learning without further fine-tuning.

\\

\cline{2-9}

&
\centering Retrieval Augmented Generation (During Generation)&
\raggedright A Stitch in Time Saves Nine: Detecting and Mitigating Hallucinations of LLMs by Validating Low-Confidence Generation \cite{varshney2023stitch} & {\color{mygreen}\checkmark}&

Article generation task, Multi-hop Questions, False Premise Questions
& Accuracy and Success & GPT-3.5,
Vicuna & Manual & Details not provided\\
\cline{3-9}

& & \raggedright A Step Closer to Comprehensive Answers: Constrained Multi-Stage Question Decomposition with Large Language Models \cite{cao2023step} & {\color{myred}$\times$}&

QA
& Recall and F1 score & PaLM, InstructGPT, GPT-3 and LLaMA2 & ChitChatQA and HotPotQA & Details not provided\\

\cline{3-9}

& & \raggedright EVER: Mitigating Hallucination in Large Language Models through Real-Time Verification and Rectification \cite{kang2023ever} & {\color{mygreen}\checkmark}&

Short-form QA, Biography
generation, and Reasoning,
& Exact match(EM),F1-score,\newline recall @ 5,\newline FACTSCORE & InstructGPT, Llama 2 7B Chat, Llama 2 13B Chat, Llama 1 65B and GPT-3.5 & HotPotQA,\newline TriviaQA,\newline ALCE-Qampari QA,bio benchmark
& $\bullet$ The paper focuses solely on enhancing text attribution to reduce hallucinations. 

$\bullet$ It relies on references, which may have inaccuracies, to support facts.\\

\cline{2-9}

&
\centering Retrieval Augmented Generation (After Generation)&
\raggedright RARR: Researching and Revising What Language Models Say, Using Language Models \cite{gao2023rarr}  & {\color{mygreen}\checkmark}& Editing for Attribution & \raggedright Attributable to Identified Sources (AIS),\newline automated metric, auto-AIS, Preservation(intent, \newline Levenshtein similarity, combined) & PaLM 540B, GPT-3,\newline LaMDA, \newline EFEC & NQ, SQA and QReCC & $\bullet$ Evaluation metrics don't cover all attribution aspects, like self-evident sentences.

$\bullet$ Preservation metrics penalize necessary revisions for severely flawed input text.

$\bullet$ RARR isn't equipped for long documents due to a lack of examples in the few-shot LLM prompts.

$\bullet$ It tends to retain unattributed claims, some of which may be hallucinations.

$\bullet$ The model is computationally intensive.\\

\cline{3-9}

& & \raggedright The Troubling Emergence of Hallucination in Large Language Models – An Extensive Definition, Quantification, and Prescriptive Remediations \cite{rawte2023troubling} & {\color{mygreen}\checkmark}&

Text Generation and QA & HVI & T5, XLNet, T0, BLOOM, Alpaca, GPT-4, OPT, Dolly, GPT-3.5, LLaMA, MPT, Vicuna, GPT-2, StableLM, GPT-3 & HILT (Own dataset) 
& $\bullet$ The paper annotated only one category per sentence for simplicity.

$\bullet$ The defined hallucination categories might not cover emerging types.

$\bullet$ The benchmark includes 15 contemporary models and might overlook recent LLM developments.\\

\cline{2-9}

&
\centering Retrieval Augmented Generation (End-to-End)&
\raggedright Retrieval-Augmented Generation for
Knowledge-Intensive NLP Tasks \cite{lewis2021RAG} & {\color{myred}$\times$}&

Open-domain QA, Abstractive QA, Jeopardy QG, Fact Verification
& BLEU-1, Q-BLEU-1, Rouge-L & T5 11B, BART & NQ, TriviaQA, WebQuestions, CuratedTrec,  MSMARCO, SearchQA, FEVER-3, FEVER-2 & Details not provided\\

\cline{2-9}
&
\centering Self-refinement through feedback and reasoning&
\raggedright  Prompting GPT-3 to Be Reliable \cite{si2022prompting}&
{\color{mygreen}\checkmark}&
\raggedright QA&
\raggedright Accuracy, Expected Calibration Error (ECE) and Brier score &DPR-BERT, GPT-3& NQ, TriviaQA,
and HotpotQA 
& 
$\bullet$ Explores four reliability facets but doesn't analyze to understand model behaviors.\\
\cline{3-9}

& & \raggedright Self-Contradictory Hallucinations of LLMs: Evaluation, Detection and Mitigation \cite{mündler2023selfcontradictory} & {\color{mygreen}\checkmark}&

Open-domain text generation,
& Self contra reduced,
Informative facts retained,
Perplexity increased & GPT-4, GPT-3.5, Llama2 70B Chat, and
Vicuna & Manual & Details not provided\\
\cline{3-9}
& & \raggedright Mind’s Mirror: Distilling Self-Evaluation Capability and
Comprehensive Thinking from Large Language Models \cite{liu2023minds}  & {\color{mygreen}\checkmark}& Short-form QA, biography
generation, and reasoning, & \raggedright Exact match (EM) and F1-score,\newline recall@5,\newline FACTSCORE & InstructGPT, \newline Llama 2 7B Chat, \newline Llama 2 13B Chat, \newline Llama 1 65B and GPT-3.5 & HotPotQA,\newline TriviaQA,\newline ALCE-Qampari QA \newline ,bio benchmark & 
$\bullet$ Experiments are conducted primarily utilizing only a single teacher model, GPT-3.5, and one student model, T5-Base.

$\bullet$ The work only evaluates their methods on three different NLP tasks.

$\bullet$ Flaws or biases in the LLMs self-evaluation mechanism may propagate to the distilled SLM.\\

\cline{3-9}

& & \raggedright Towards Mitigating Hallucination in Large Language Models
via Self-Reflection \cite{ji2023mitigating} & {\color{mygreen}\checkmark}&

Medical generative QA & unigram F1 and ROUGE-L & Vicuna,\newline Alpaca-
LoRA,\newline GPT-3.5, \newline MedAlpaca,\newline Robin-medical & PubMedQA,
\newline MedQuAD,
\newline MEDIQA\newline2019,
\newline LiveMedQA\newline2017,
MASH-QA
& $\bullet$ The study is limited to English medical queries, limiting the generalizability to other languages, domains, and modalities.\\

\cline{3-9}

& & \raggedright On What Basis? Predicting Text Preference Via Structured Comparative Reasoning \cite{yan2023basis} & {\color{mygreen}\checkmark}&

Summarization, retrieval, and automatic rating & Accuracy & GPT-3.5 and GPT-4 & TL;DR, RLAIF-HH and TREC News
& $\bullet$ The evaluation is conducted on a sample set of datasets. 

$\bullet$ Consistency measurement uses approximate metrics rather than more rigorous schemes.\\

\cline{3-9}

& & \raggedright DRESS: Instructing Large Vision-Language Models to
Align and Interact with Humans via Natural Language Feedback \cite{chen2023dress} & {\color{mygreen}\checkmark}&

Visual QA & Helpfulness, honesty and harmlessness & BLIP-2 with T5-XXL, LLaVA with LLaMA-13B, LLaVA-HF with Vicuna,
InstructBLIP with Vicuna, MiniGPT-4 with Vicuna, mPLUG-Owl with LLaMA-7B & BLIP, CC3M, CC12M, SBU, LLaVA and VLSafe (Own dataset) & Details not provided \\

\cline{3-9}

& & \raggedright The Knowledge Alignment Problem: Bridging Human and External
Knowledge for Large Language Models \cite{zhang2023knowledge} & {\color{mygreen}\checkmark}&

QA& G-EVAL: Gold Answer Coverage, Hallucination, Accepted & GPT-3.5, & FuzzyQA & 
$\bullet$ The additional clarification steps increase the computational load and time consumption.
\\

\cline{3-9}

& & \raggedright Chain-of-Verification Reduces Hallucination in Large Language Models \cite{dhuliawala2023chainofverification} & {\color{mygreen}\checkmark}&

QA& Precision & Llama 65B, & QUEST,\newline MultiSpan-QA & $\bullet$ The work only addresses hallucinations in the form of directly stated factual inaccuracies.

$\bullet$ Computational cost is increased due to generating verification statements and additional model deliberation.\\

\cline{3-9}

& & \raggedright Chain of Natural Language Inference for Reducing
Large Language Model Ungrounded Hallucinations \cite{lei2023chain} & {\color{mygreen}\checkmark}&

Summarization and question answering & F1,Rouge- 1,2, L, Bleu-4, BertScore, FactCC and
AlignScore-Large & GPT-3.5
and GPT-4 & HaluEVAL,
FactCC,
SummEval, QAGS-Xsum, QAGS-CNNDM & Details not provided \\

\cline{2-9}

&
\centering Prompt Tuning &
\raggedright  UPRISE: Universal Prompt Retrieval for Improving Zero-Shot Evaluation \cite{cheng2023uprise}&
{\color{mygreen}\checkmark}&
\raggedright QA&
\raggedright Accuracy & \raggedright Gpt-3.5, Gpt-Neo-2.7B, BLOOM-7.1B, OPT-66B, GPT3 & TruthfulQA,
FEVER2.0,
the scientific spilt of Covid-19 & $\bullet$ It displays limited impact on tasks that are directly formulated as language modeling, such as coreference resolution and commonsense reasoning.\\
\cline{3-9}

& & \raggedright Teaching Language Models to Hallucinate Less with Synthetic Tasks \cite{jones2023teaching} & {\color{mygreen}\checkmark}&

Search-and-retrieve, meeting summarization, and clinical report generation
& ROUGE-1, ROUGE-2, and ROUGE-L & \raggedright Vicuna, v1.1 13B, GPT-4 & \raggedright MS MARCO, QMSum, ACI-Bench & $\bullet$ It requires designing a synthetic task, and reduces hallucination on some models more than others. \\
\cline{1-9}

\multirow{14}{*}[-20ex]\textbf{Developing Models} &
\centering Introducing new decoding strategy&
\raggedright  Trusting Your Evidence:
Hallucinate Less with Context-aware Decoding \cite{shi2023trusting}&
{\color{myred}$\times$}&
\raggedright Summarization,
Knowledge Conflicts&
\raggedright ROUGE-L, BERT-Precision, FactKB &OPT(13B and 30B), GPT-Neo (2.7B and 20B), LLaMA (13B and 30B) and FLAN-T5 (XL 3B and XXL 11B)&CNN-DM, XSUM, MemoTrap,
NQ-Swap & Details not provided\\
\cline{3-9}
& & \raggedright DOLA: Decoding by Contrasting Layers Improves Factuality in Large Language Models \cite{chuang2023dola} & {\color{mygreen}\checkmark}&

Multiple choices tasks and open-ended generation
& GPT-4 automatic evaluation & LLaMA-(7B, 13B, 33B, 65B) and GPT4 & TruthfulQA,\newline FACTOR,\newline StrategyQA,\newline GSM8K, \newline
Vicuna QA & $\bullet$ The work doesn't explore performance in other dimensions like instruction following or learning from human feedback.

$\bullet$ It relies on existing architecture and pre-trained parameters, omitting the utilization of human labels or factual knowledge bases for fine-tuning, thereby limiting potential improvements.

$\bullet$ This method solely relies on the model's internal knowledge and lacks external retrieval modules, which may result in an inability to correct misinformation acquired during training.
\\
\cline{3-9}
& & \raggedright Inference-Time Intervention:
Eliciting Truthful Answers from a Language Model \cite{li2023inference}  & {\color{mygreen}\checkmark}& QA & \raggedright Attributable to Identified Sources (AIS),\newline True*Informative (\%), True (\%), MC acc. (\%), CE (pre-training loss) and KL (divergence between next-token distributions pre- and post-intervention) & LLaMA, Alpaca and Vicuna & TruthfulQA & Details not provided \\

\cline{2-9}

&
\centering Utilization of Knowledge Graph&
\raggedright RHO: Reducing Hallucination in Open-domain Dialogues with
Knowledge Grounding \cite{ji-etal-2023-rho} & {\color{myred}$\times$}&

Open-domain dialogue response generation
& BLEU, ROUGE-L & GPT-2, BART, GPT-3.5 & OpenDialKG & $\bullet$ The model identifies statistical patterns and quantitative links among variables but cannot perceive qualitative relationships like causality, hierarchy, and other abstractions.\\
\cline{3-9}

& & \raggedright FLEEK: Factual Error Detection and Correction with Evidence Retrieved from External Knowledge \cite{bayat2023fleek} & {\color{mygreen}\checkmark}&

Fact verification and Fact Revision,
& Accuracy, precision, recall, and F1 score & Vicuna and GPT-3 & BenchLLM and BenchText & $\bullet$ The current system relies on the initial set of responses generated by LLMs to execute tasks.

$\bullet$ The experiments presented are based on small-scale datasets.\\
\cline{2-9}

&
\centering Introducing faithfulness-based loss function &
\raggedright Information-Theoretic Text Hallucination Reduction for Video-grounded Dialogue \cite{yoon-etal-2022-information}  & {\color{myred}$\times$}& Video-grounded Dialogues & \raggedright BLEU, METEOR, ROUGE-L, CIDEr & T5 &AVSD@ \newline DSTC7, \newline AVSD@ \newline DSTC8 & $\bullet$ It requires pre-training each language model in a two-stage training mechanism to mitigate text hallucination.\\

\cline{3-9}

& & \raggedright Detecting and Mitigating Hallucinations in Multilingual Summarisation \cite{qiu2023detecting} & {\color{mygreen}\checkmark}&

Multilingual Summarisation & faithfulness,\newline DAE, QAFactEval,ENFS\%,
EntFA,
ROUGE-1/2/L scores & BLOOMZ-P3-7.1B, Vicuna, Phoenix-7B, & XL-Sum & $\bullet$ It uses machine translation to construct training data, which may limit feasibility for other languages. Translation errors may also limit metric quality.

$\bullet$ The weighted-loss approach has inconsistent gains in faithfulness across languages. \\

\cline{2-9}

&
\centering Supervised finetuning&
\raggedright  HALO: Estimation and Reduction of Hallucinations in Open-Source Weak Large Language Models \cite{elaraby2023halo}&
{\color{mygreen}\checkmark}&
\raggedright QA &
\raggedright HaloCheck (Own metric) &BLOOM 7B, GPT-4& Manual & $\bullet$ The study includes only one example of a weak open-source LLM (BLOOM7B) and concentrated solely on the NBA domain for analysis.

$\bullet$ Relies on automatically generated questions.\\
\cline{3-9}

& & \raggedright Hallucination Augmented Recitations for Language Models \cite{köksal2023hallucination} & {\color{myred}$\times$}& 

QA,
& F1-score & T5 & CFTriviaQA & Details not provided\\
\cline{3-9}
& & \raggedright Fine-tuning Language Models for Factuality \cite{tian2023finetuning}  & {\color{mygreen}\checkmark}& Biography generation and Medical QA & \raggedright FactScore & Llama1 and Llama2-chat & Manual & Details not provided\\

\cline{3-9}

& & \raggedright Dial BEINFO for Faithfulness: Improving Factuality of
Information-Seeking Dialogue via Behavioural Fine-Tuning \cite{razumovskaia2023textitdial} & {\color{mygreen}\checkmark}&

QA & BLEU, ROUGE, BERTScore and Precision & Flan-T5 (Base, Large and XL) & FaithDial, TopiOCQA and DoQA
& $\bullet$ The work concentrates on models with a parameter limit of up to 3B.

$\bullet$ It addresses the reduction of LLM hallucinations in information-seeking dialogue without intervening in the knowledge retrieval component.\\

\cline{3-9}

& & \raggedright R-Tuning: Teaching Large Language Models to
Refuse Unknown Questions \cite{zhang2023r} & {\color{myred}$\times$}&

QA & Accuracy, Average Precision (AP) & OpenLLaMA-3B, LLaMA-7B and LLaMA-13B  & ParaRel, MMLU, WiCE, HotpotQA and FEVER & Details not provided\\
\cline{3-9}

& & \raggedright Think While You Write Hypothesis Verification Promotes Faithful Knowledge-to-Text Generation \cite{qiu2023think} & {\color{mygreen}\checkmark}&
Hypothesis Verification & FactKB, BLEU, METEOR, BERTScore & BART-large and T5-large & FATE (Own novel dataset), WebNLG, TekGen and GenWiki & \raggedright
$\bullet$ The TWEAK decoding strategy increases computational cost during inference compared to baseline approaches like beam search.

$\bullet$ The approach has only undergone testing in the English language.

\label{tab:table_1}

\end{longtable}
\clearpage
\twocolumn

\end{document}